\documentclass[letterpaper, 10 pt, conference]{ieeeconf}  

\IEEEoverridecommandlockouts                              

\usepackage{amsmath,amssymb,amsfonts}
\usepackage{algorithmic}
\usepackage{graphicx}
\usepackage{textcomp}
\def\BibTeX{{\rm B\kern-.05em{\sc i\kern-.025em b}\kern-.08em
    T\kern-.1667em\lower.7ex\hbox{E}\kern-.125emX}}
    
\usepackage{graphicx,epstopdf}

\usepackage{balance}
\usepackage{url}
\usepackage{hyperref}
\hypersetup{
colorlinks=true,
linkcolor=blue,
filecolor=blue,      
urlcolor=blue,
citecolor=cyan,
}
\usepackage{amssymb,amsthm}
\usepackage{amsmath}
\usepackage{amsfonts}
\usepackage{mathrsfs}
\usepackage{tabularx} 
\usepackage[linesnumbered,boxed,ruled,commentsnumbered]{algorithm2e}

\usepackage{subfigure}
\usepackage{acronym}
\usepackage{todonotes}
\usepackage{color}
\usepackage{mathtools}
\usepackage{booktabs}
\usepackage{multirow}

\newif\ifuseboldmathops
\newif\ifuseittextabbrevs

\ifuseittextabbrevs

\else

\fi

\ifuseboldmathops

\else

\fi

\ifuseboldmathops

\else

\fi

\ifuseboldmathops

\else

\fi

\ifuseboldmathops

\else

\fi

\acrodef{dm}[DM]{Direct Modeling}
\acrodef{mbb}[MBB]{Moving Block Bootstrap}
\acrodef{cm}[CM]{Combined Method}
\acrodef{bev}[BEV]{Bird's-Eye-View}
\acrodef{kl}[KL]{Kullback–Leibler}
\acrodef{ap}[AP]{Average Precision}
\acrodef{iou}[IoU]{Intersection-over-Union}
\acrodef{lb}[LB]{Lower-bound}
\acrodef{ub}[UB]{Upper-bound}
\acrodef{dn}[DN]{DiscoNet}
\acrodef{nll}[NLL]{Negative Log Likelihood}
\acrodef{ece}[ECE]{Expected Calibration Error}
\acrodef{mdm}[Double-M Quantification]{direct-Modeling Moving-block bootstrap Quantification}
\newcommand{\mdm}{Double-M Quantification (direct-Modeling Moving-block bootstrap Quantification)}

\theoremstyle{definition}
 
 \newtheorem{example}{Example}

\begin{document}

\author{
{Sanbao Su} \and {Yiming Li} \and {Sihong He} \and {Songyang Han} \and {Chen Feng} \and {Caiwen Ding} \and {Fei Miao} 
\thanks{This work was supported by NSF 1932250, 1952096,  2047354, and 2121391 grants. Sanbao~Su, Sihong~He, Songyang~Han, Caiwen~Ding, and Fei~Miao are with the Department of Computer Science and Engineering, University of Connecticut, Storrs Mansfield, CT, USA 06268. Email: \{sanbao.su, sihong.he, songyang.han, caiwen.ding, fei.miao \}@uconn.edu. Yiming Li and Chen Feng are with Tandon School of Engineering, New York University, Brooklyn, NY, USA 11201. Email:\{yimingli, cfeng \}@nyu.edu.}
}


\title{\LARGE \bf
Uncertainty Quantification of Collaborative Detection for Self-Driving
}

\maketitle

\thispagestyle{empty}
\pagestyle{empty}

\begin{abstract}
Sharing information between connected and autonomous vehicles (CAVs) fundamentally improves the performance of collaborative object detection for self-driving. However, CAVs still have uncertainties on object detection due to practical challenges, which will affect the later modules in self-driving such as planning and control. Hence, uncertainty quantification is crucial for safety-critical systems such as CAVs. Our work is the first to estimate the uncertainty of collaborative object detection. We propose a novel uncertainty quantification method, called Double-M Quantification, which tailors a moving block bootstrap (MBB) algorithm with direct modeling of the multivariant Gaussian distribution of each corner of the bounding box. Our method captures both the epistemic uncertainty and aleatoric uncertainty with one inference pass based on the offline Double-M training process. And it can be used with different collaborative object detectors. Through experiments on the comprehensive collaborative perception dataset, we show that our Double-M method achieves more than 4$\times$ improvement on uncertainty score and more than 3\% accuracy improvement, compared with the state-of-the-art uncertainty quantification methods. Our code is public on \url{ https://coperception.github.io/double-m-quantification/}. 
\end{abstract}

\section{introduction}
\label{sec:intro}
Multi-agent collaborative object detection has been proposed to leverage the viewpoints of other agents to improve the detection accuracy compared with the individual viewpoint~\cite{li2022v2x}. Recent research has shown the effectiveness of early, late, and intermediate fusion of collaborative detection, which respectively transmits raw data,  output bounding boxes, and intermediate features~\cite{li2021learning,Chen2019CooperCP,arnold2020cooperative,chen2022model,li2022multi}, and the improved collaborative object detection results will benefit the self-driving decisions of connected and autonomous vehicles (CAVs)~\cite{han2022stable}. However, CAVs may still have uncertainties on object detection due to out-of-distribution objects, sensor measurement noise, or poor weather~\cite{liu2021robust,kothandaraman2021ss,feng2021review}. Even a slightly false detection can lead the autonomous vehicle's driving policy to a completely different action~\cite{huang2017adversarial,lin2017tactics}. For example, miss-detected paint on the road surface can confuse the lane-following policy and cause potential accidents~\cite{kurakin2016adversarial}. Therefore, it is crucial to quantify the uncertainty of object detection for safety-critical systems such as CAVs.

Various uncertainty quantification methods have been proposed for object detection~\cite{feng2021review,hall2020probabilistic,harakeh2021estimating,chiu2021probabilistic}.
Uncertainty resources can be decomposed into aleatoric and epistemic uncertainty~\cite{hullermeier2021aleatoric,kendall2017uncertainties}. Direct modeling methods~\cite{meyer2020learning,he2019bounding,he2020deep} focus on the aleatoric uncertainty (or data uncertainty) which represents inherent measurement noises from the sensor. Monte-Carlo dropout~\cite{gal2017concrete,miller2018dropout} and deep ensemble~\cite{lakshminarayanan2017simple} methods focus on the epistemic uncertainty (or model uncertainty) which reflects the degree of uncertainty that a model describes an observed dataset with its parameters.
However, none of the above methods have investigated the uncertainty quantification of collaborative object detection.

\begin{figure}[t]
    \centering    
    \includegraphics[width=0.48\textwidth]{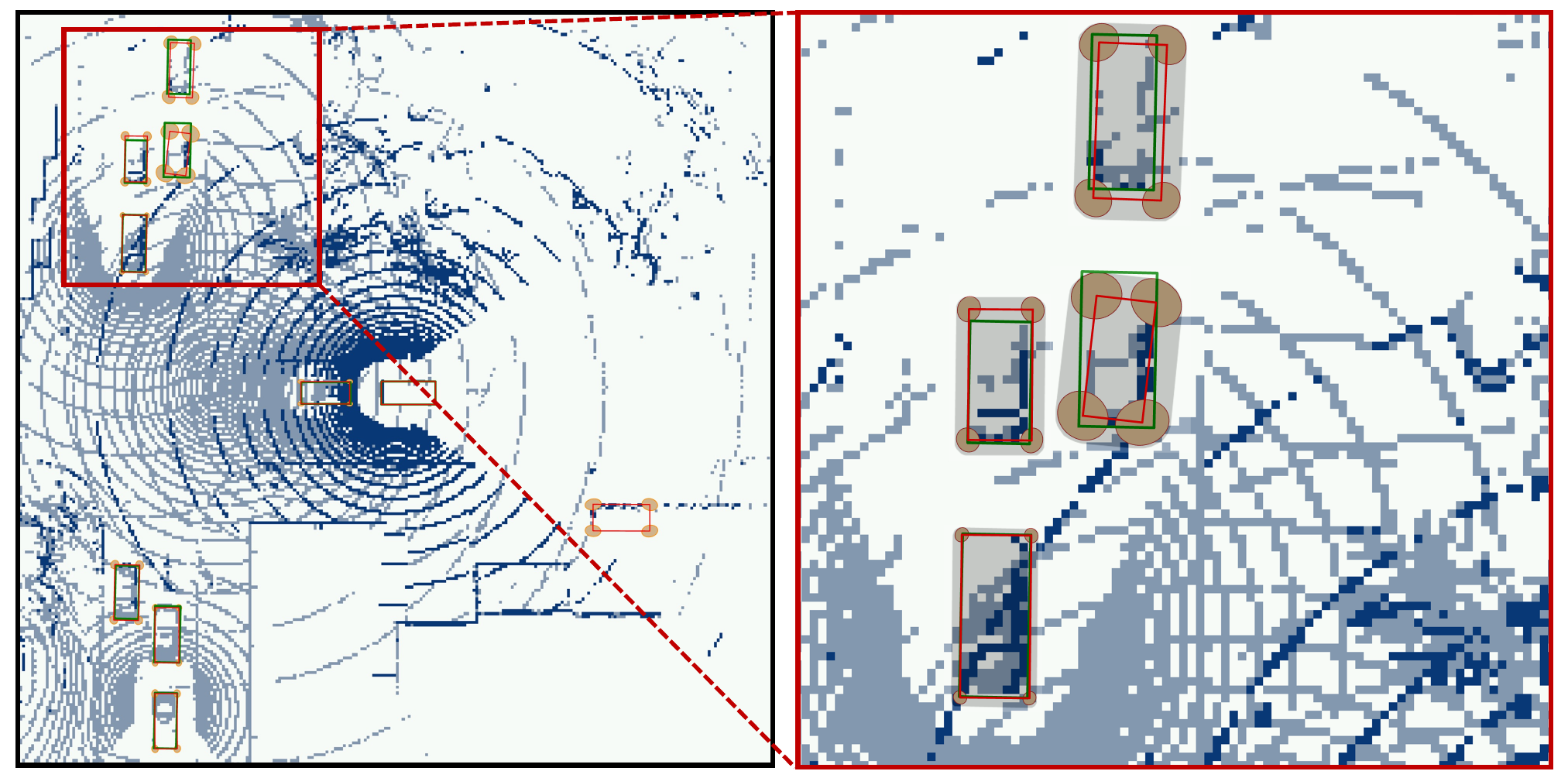}
    \vspace{-8mm}
    \caption{ \textbf{Left} figure shows detection results of intermediate collaboration in bird's eye view (BEV), and \textbf{right} figure zooms on a specific part to show the robust range of two detections. Red boxes are predictions, and green boxes are ground truth. The orange ellipse denotes the covariance of each corner. The shadow convex hull shows the uncertainty set of the detected object. The shadow convex hull covers the green bounding box in most cases, which helps the later modules in self-driving tasks, such as trajectory prediction with uncertainty propogation~\cite{boris2022propagating} and robust planning and control~\cite{SurveyPlan_16, motion_icra14}. With our Double-M Quantification method, detected objects with low accuracy tend to have large uncertainties.
    }
    \label{fig:vis}
    \vspace{-5mm}
\end{figure}

In this paper, we propose a novel uncertainty quantification method for collaborative  object detection, called \mdm,
that only needs one inference pass to capture both the epistemic and aleatoric uncertainties.
The constructed uncertainty set for each detected object by our method helps the later modules in self-driving tasks, such as trajectory prediction with uncertainty propogation~\cite{boris2022propagating} and robust planning and control~\cite{SurveyPlan_16, motion_icra14}. From Fig.~\ref{fig:vis}, we can see that with our uncertainty quantification method, detected objects with low accuracy tend to have large uncertainties, and the constructed uncertainty set covers the  ground-truth bounding box in most cases. Compared with the state-of-the-arts~\cite{he2019bounding,he2020deep}, 
our Double-M Quantification method achieves up to 4$\times$ improvement on uncertainty score and up to 3.04\% accuracy improvement on the comprehensive collaborative  perception dataset, V2X-SIM~\cite{li2022v2x}.

The main contributions of this work are as follows:
\begin{enumerate}
    \item To the best of our knowledge, our proposed Double-M Quantification is the first attempt to estimate the uncertainty of collaborative object detection. 
    Our method tailors a moving block bootstrap algorithm to estimate both the epistemic and aleatoric uncertainties in one inference pass. 
    \item We design a novel representation format of the bounding box uncertainty in the direct modeling component to estimate the aleatoric uncertainty. We consider each corner of the bounding box as one independent multivariant Gaussian distribution and the covariance matrix for each corner is estimated from one output header, while the existing literature mainly assumes a univariant Gaussian distribution for each dimension of each corner or a high dimensional Gaussian distribution for all corners. 
    \item  We validate the advantages of the proposed methodology based on V2X-SIM~\cite{li2022v2x} and show that our Double-M Quantification method reduces the uncertainty and improves the accuracy. The results also validate that sharing intermediate feature information between CAVs is beneficial for the system in both improving accuracy and reducing uncertainty. 
\end{enumerate}

\section{Related Work}
\label{sec:relatedwork}

\paragraph{Collaborative Object Detection} Collaborative object detection, in which multiple agents collectively perceive a scene via communication, is able to address several dilemmas in individual object detection~\cite{li2022v2x,cai2022analyzing}. Compared to individual detection, multi-agent collaboration introduces more viewpoints to solve the long-range data sparsity and severe occlusions.
The pioneer collaborative detectors employ early collaboration which shares raw data~\cite{Chen2019CooperCP} or late collaboration which shares output bounding boxes~\cite{arnold2020cooperative}. To further improve the performance-bandwidth trade-off, recent research proposes intermediate collaboration which shares intermediate feature representations from a neural network. Various intermediate collaboration strategies have been developed such as neural message passing~\cite{wang2020v2vnet}, knowledge distillation~\cite{li2021learning}, and attention~\cite{xu2022v2x,xu2022bridging}. However, existing works only focus on improving the performance of collaborative detection, no existing work investigates uncertainty quantification of collaborative object detection. 

\paragraph{Uncertainty Quantification on Object Detection} 
Different types of uncertainty quantification methods for object detection have been proposed. For epistemic uncertainty, the Monte-Carlo dropout method utilizes the dropout-based neural network training to perform approximated inference in Bayesian neural networks~\cite{miller2018dropout}. The deep ensembles method estimates probability distribution by an ensemble of networks with the same architecture and different parameters~\cite{lakshminarayanan2017simple,lyu2020probabilistic,ovadia2019can}. Both methods require multiple runs of inference, which makes them infeasible for real-time critical tasks with high computational costs such as collaborative objection detection. Moreover, they do not consider time series properties in the dataset, which is one important characteristic of the autonomous driving dataset. In contrast, our uncertainty quantification method overcomes these problems by tailoring a moving block bootstrap~\cite{lahiri1999theoretical} (MBB, an effective algorithm for time series analysis) algorithm and quantifies the uncertainty of collaborative object detection in one inference pass.

The direct modeling (DM) method is designed to estimate aleatoric uncertainty. The main steps of DM are~\cite{feng2021review}: a) select one object detector; b) set a certainty probability distribution on outputs of the detector and design the corresponding loss function; c) add extra regression layers to predict the covariance; d) train the modified detector. The work~\cite{he2019bounding} proposes the DM method for image object detection, which assumes that the distribution of each bounding box variable is a single-variate Gaussian distribution and introduces one additional layer to estimate the variance of the bounding box. \cite{he2020deep} proposes the DM method with a high dimensional multivariate Gaussian distribution. DM methods for point cloud object detection~\cite{meyer2020learning,meyer2019lasernet} have been proposed. Methods with both DM and MC dropout to estimate aleatoric and epistemic uncertainties in object detection have also been investigated~\cite{kendall2017uncertainties,feng2019leveraging,feng2018towards}. 

All the above works only focus on individual object detection. How to quantify the aleatoric and epistemic uncertainties in collaborative object detection remains challenging. In our work, we tailor an MBB-based algorithm process to estimate both aleatoric and epistemic uncertainties of collaborative object detection, with an independent multivariate Gaussian distribution assumption for each corner of the bounding box to represent the uncertainty.



\section{Uncertainty Quantification Approach}
\label{sec:approach}

\begin{figure*}[t]
    \centering    
    \includegraphics[width=0.95\textwidth]{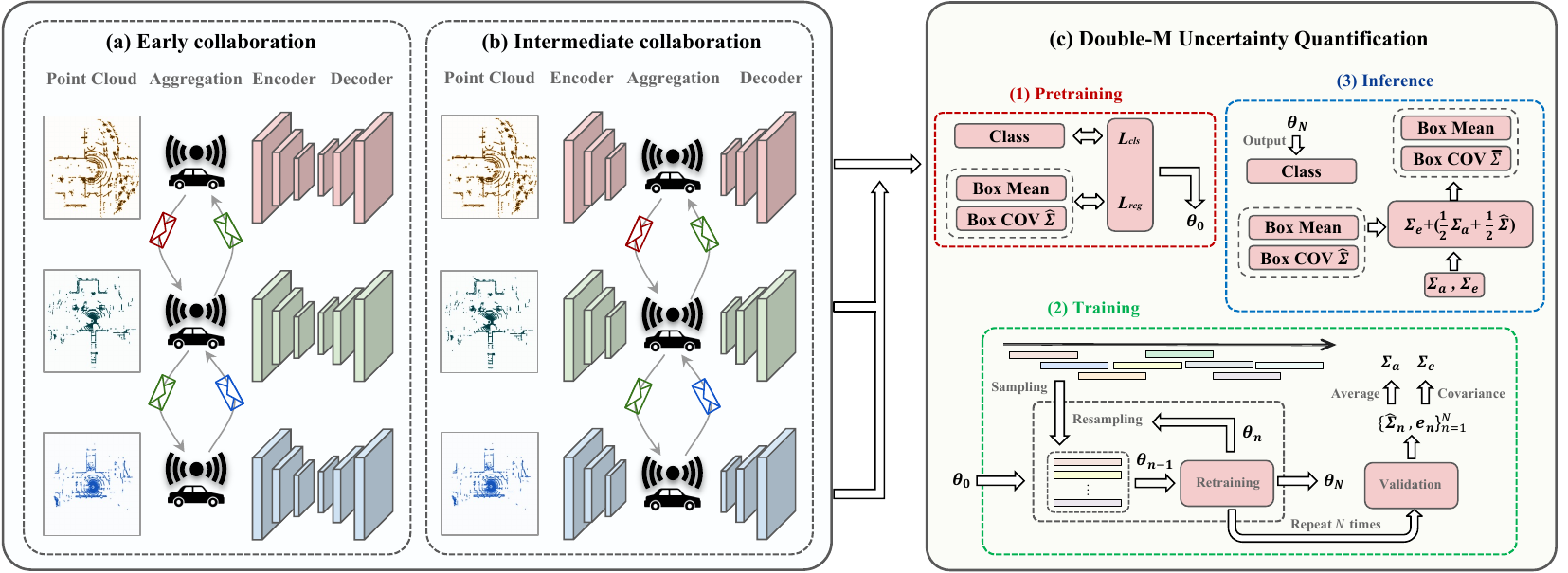}
    \caption{Overview of our Double-M Quantification method on collaborative object detection. \textbf{(a) Early collaboration} shares raw point cloud with other agents, and \textbf{(b) intermediate collaboration} shares intermediate feature representations with other agents. \textbf{(c) Double-M Quantification method} estimates the multi-variate Gaussian distribution of each corner.
    Our Double-M Quantification method can be used on different collaborative object detection. During the training stage, Double-M Quantification tailors a moving block bootstrapping algorithm to get the final model parameter, $\Sigma_{a}$ as the average aleatoric uncertainty of the validation dataset and $\Sigma_{e}$ as the covariance of all residual vectors for epistemic uncertainty. During the inference stage, combine $\Sigma_a$, $\Sigma_e$ and the predicted covariance matrix $\hat{\Sigma}$ from the object detector to compute the covariance matrix $\bar{\Sigma}=\Sigma_{e} + (\frac{1}{2}\Sigma_{a} + \frac{1}{2}\hat{\Sigma})$ of the distribution.
    }
    \label{fig:main}
    \vspace{-20pt}
\end{figure*}

In this section, we first define the problem of uncertainty quantification for collaborative object detection. Then we describe the overview structure of our novel \mdm \ method as in Fig.~\ref{fig:main}, followed by the detailed algorithm process. Finally, we define our loss function of the neural network model. One major novelty is the first to tailor a moving block bootstrapping~\cite{lahiri1999theoretical} (MBB) algorithm to address the uncertainty quantification challenge of collaborative object detection, and estimate the epistemic and aleatoric uncertainty with one inference pass in addition to the offline training process. The algorithm does not rely on a specific neural network model or structure and can be used with different collaborative object detectors such as DiscoNet~\cite{li2022v2x}. The corresponding loss function considers both the prediction accuracy and covariance as metrics. 

\subsection{Problem Description}

We use $X$ to represent the point cloud data. For the deterministic collaborative object detection, the agent $\alpha \in \{1,\ldots,N \}$ creates a local binary~\ac{bev} map $\mathcal{M} \in \{0, 1\}^{W_m \times L_m \times H_m}$ from $X$, where $N$ is the number of agents, and $W_m$, $L_m$, and $H_m$ indicate the width, length, and height of the local~\ac{bev} map respectively~\cite{wu2020motionnet}. As shown in Fig.~\ref{fig:main} (b), we consider homogeneous collaborative object detection, in which multiple agents share the same detector $f_\theta$ which consists of an encoder denoted by $\mathcal{E}$, and an aggregator as well as a decoder that is collectively represented by $\mathcal{D}$, where $\theta$ denotes the parameters of $f$. $\mathcal{E}$ compresses the~\ac{bev} map into an intermediate feature map $\mathbf{F}_\alpha \in \mathbb{R}^{\frac{W_m}{K_m} \times \frac{L_m}{K_m} \times F_m} = \mathcal{E}(\mathcal{M})$, where $K_m$ is the spatial downsampling scale, and $F_m$ is the feature dimension. Afterwards, $\mathbf{F}_\alpha$ is sent to other agents, and the agent $\alpha$ will use $\mathcal{D}$ to generate a set of bounding boxes $\hat{Y} = \mathcal{D}(\mathbf{F}_\alpha, \{ \mathbf{F}_\beta \}_{\beta \neq \alpha})$. We also consider early collaboration which shares raw point cloud as shown in Fig.~\ref{fig:main} (a), we omit its notations for simplicity because the only thing that sets it apart from the intermediate collaboration is the timing of  sharing information.

In each point cloud data $X$, there are $J$ objects. For each object $j \in \{1, \ldots, J \}$, we propose to predict $I$ corners of the bounding box. Each corner $i \in \{1, \ldots, I\}$ is represented by a $D$-dimensional vector in the~\ac{bev} map. The set of ground-truth bounding boxes $Y$ is represented as $ Y = \{ c_j, \{y_{ij}\}_{i=1}^I \}_{j=1}^J$ where $c$ is the classification label and $y_{ij} \in \mathbb{R}^D, \forall (i,j)$. The set of predicted bounding box $\hat{Y}$ is represented as $\hat{Y} = \{ \hat{p}_j,  \{\hat{y}_{ij}, \hat{\Sigma}_{ij}\}_{i=1}^I \}_{j=1}^J $, where $\hat{p}$ is the predicted classification probability, and each corner of the bounding box is modeled as a multivariate Gaussian distribution with $\hat{y}_{ij} \in \mathbb{R}^D$ as the predicted mean and $\hat{\Sigma}_{ij}$ as the $D \times D$ covariance matrix coordinate of corner $i$ of object $j$. Here we assume the probability distribution of each corner of the bounding box is independent. During training, the neural network parameters of the encoder $\mathcal{E}$ , aggregator and decoder $\mathcal{D}$ are jointly learned by minimizing the detection loss $\mathcal{L}_{det}(Y, \hat{Y})$ that includes a classification loss and a regression loss that considers both prediction accuracy and uncertainty.



\subsection{Solution Overview}

We design a novel uncertainty quantification method called \ac{mdm} to estimate epistemic and aleatoric uncertainties by tailoring an MBB algorithm with the DM method. The overview of \ac{mdm} on collaborative object detection is shown in Fig.~\ref{fig:main}. During the training stage, we train the object detector on resampled moving blocks. After $N$ bootstraps, we get the object detector $f_{\hat{\theta}}$ where $\hat{\theta}$ is the final model parameter, compute $\Sigma_a$ as the average aleatoric uncertainty of the validation dataset, and compute $\Sigma_e$ as the covariance of all residual vectors for epistemic uncertainty. During the inference stage, with input point cloud $X$, we combine $\Sigma_a$, $\Sigma_e$ and the predicted covariance matrix $\hat{\Sigma}_{ij}$ from $f_{\hat{\theta}}(X)$ to compute the covariance matrix $\bar{\Sigma}_{ij}=\Sigma_{e} + (\frac{1}{2}\Sigma_{a} + \frac{1}{2}\hat{\Sigma}_{ij})$ of the multivariate Gaussian distribution. In the following subsection~\ref{sec:doubleM}, we propose our novel design, the main contribution of this work, the Double-M Quantification algorithm. Then, in subsection~\ref{sec:direct_modeling}, we introduce the loss function for our method.


\subsection{Double-M Quantification}
\label{sec:doubleM}

Monte-Carlo dropout \cite{miller2018dropout} and deep ensembles \cite{lakshminarayanan2017simple} have been proposed to estimate the epistemic uncertainty. However, none of them consider the time-series features in the dataset while the temporal features are important for CAVs. We design a novel uncertainty quantification method called \ac{mdm} to estimate epistemic and aleatoric uncertainties when considering the temporal features in the dataset. In particular, our design tailors a moving block bootstrapping~\cite{lahiri1999theoretical} process on time-series data that captures the autocorrelation within the data by sampling data from the constructed data blocks in the training process.

\setlength{\textfloatsep}{0pt}
\begin{algorithm}
\caption{Double-M Quantification - Training}
\label{alg:mbb}
 \KwData{The training dataset $\mathbf{D}_{K} = \{ (X_{k}, Y_{k}) \}_{i=k}^{K}$ with $K$ frames, the validation dataset $\mathbf{V}_{K'} = \{ (X_{k}, Y_{k}) \}_{k=1}^{K'}$ with $K'$ frames, the block length $l$, number of bootstraps $N$, the loss function $\mathcal{L}_{det}$ in Eq~\eqref{eq:L}, the collaborative object detector $f_\theta$}
 
 \KwResult{$\hat{\theta}$,  $\Sigma_{e}$, $\Sigma_{a}$}

 
 $\theta_0 \leftarrow$ argmin $\mathcal{L}_{det}(\mathbf{D}_{K}, \theta)$\\
 Construct the overlapping moving-block collection $\mathbf{B} = \{\mathcal{B}_b\}_{b=1}^{K-l+1}$ from the time-series training dataset $\mathbf{D}_K$ with each block $\mathcal{B}_b = \{ (X_{b}, Y_{b}), ..., (X_{b + l - 1}, Y_{b+ l - 1}) \}, b = 1, 2, ..., K - l + 1$. \label{alg:MBB-scene} \\
 
 \For{$n$ from $1$ to $N$}{\label{alg:MBB-boot-it-begin}
 Sample $M = \lfloor K/l \rfloor$ blocks with replacement from $\mathbf{B}$ and get a sampled dataset $\mathbf{S}_n = \{\mathcal{B}_{b_m}\}_{m=1}^{M}$ where $b_m$ are iid uniform random variables on $\{1, 2, ..., K-l+1\}$. \\ \label{alg:MBB-sample}
 Update $\theta_n \leftarrow $ argmin $\mathcal{L}_{det}(\mathbf{S}_n, \theta_{n-1})$\\ \label{alg:MBB-train}
 $f_{\theta_n}(\mathbf{V}_{K'}) = \{ \{\hat{p}_{jk}, \{\hat{y}_{ijk}, \hat{\Sigma}_{ijk}\}_{i=1}^I \}_{j=1}^J\}_{k=1}^{K'}$\\ \label{alg:MBB-test}
  Compute the residual vector $e_{ijk} = y_{ijk} - \hat{y}_{ijk}$ $\forall i\in[1,I], j\in[1,J], k\in[1,K']$ \label{alg:MBB-residual-error} \\
}\label{alg:MBB-boot-it-end}
$\hat{\theta} \leftarrow \theta_{N}$\\
Compute $\Sigma_{a}$ as the average aleatoric uncertainty of the validation dataset\\ 
Compute $\Sigma_{e}$ as the covariance of all residual vectors for epistemic uncertainty \\
\end{algorithm}


We show the training stage of Double-M Quantification method in Algorithm~\ref{alg:mbb}. We first initialize $\theta$, the parameters of a collaborative object detector and pretrain the model using the training data set. Then we construct the constant-length time-series block set $\mathbf{B}$ from the time-series training dataset $\mathbf{D}_K$ which contains $K$ frames. Notice that the block set $\mathbf{B}$ keeps the temporal characteristic (see Line~\ref{alg:MBB-scene}) by maintaining the order of frames within the same block. Then in every iteration, we retrain the model using a sampled dataset that contains $M$ blocks sampled with replacement and uniform random probability from the block set $\mathbf{B}$ (see Lines~\ref{alg:MBB-sample}--\ref{alg:MBB-train}). The sampled dataset still contains about $K$ frames since $M=\lfloor K/l \rfloor$. The floor function $\lfloor \cdot \rfloor$ is to make sure $M$ is an integer. At the final step in each training iteration $n$, we test the retained model $f_{\theta_n}$ on the validation dataset $\mathbf{V}_{K'}$ (see Line~\ref{alg:MBB-test}) and compute the residual vector as the difference between the ground-truth vector $y_{ijk}$ and the predicted mean vector $\hat{y}_{ijk}$, $\forall i\in[1,I], j\in[1,J], k\in[1,K']$ (see Line~\ref{alg:MBB-residual-error}). After $N$ iterations, we get the final model parameters $\theta_N$ as $\hat{\theta}$ to predict the covariance by model $f_{\hat{\theta}}$. Other than the final trained model, we estimate both the aleatoric and epistemic uncertainties by using the residuals and predicted covariance matrics of the validation dataset. We first estimate the aleatoric uncertainty by computing $\Sigma_{a}$, the mean of all  predicted covariance matrices. To estimate the epistemic uncertainty, we then compute the covariance matrix of all residual vectors, which is denoted by $\Sigma_{e}$. On the one hand, our Double-M Quantification method provides the bagging aleatoric uncertainty estimates through the aggregation over multiple models from the $N$ iterations on the validation dataset. On the other hand, it approximates the distribution of errors from the residuals so that we can quantify the epistemic uncertainty.

\begin{algorithm}
\caption{Double-M Quantification - Inference}
\label{alg:mbb-inf}
\KwData{$\Sigma_{e}$, $\Sigma_{a}$, input point cloud $X$, the trained collaborative object detector $f_{\hat{\theta}}$}
\KwResult{$\bar{Y}$}
$f_{\hat{\theta}}(X) = \hat{Y} = \{ \{\hat{p}_{j}, \{\hat{y}_{ij}, \hat{\Sigma}_{ij}\}_{i=1}^I \}_{j=1}^J\}$ \\
\For{j from 1 to J}{\label{alg:mbb-inf-it-start}
    \For{i from 1 to I}{
    $\bar{\Sigma}_{ij} = \Sigma_{e} + (\frac{1}{2}\Sigma_{a} + \frac{1}{2}\hat{\Sigma}_{ij})$
    }
}\label{alg:mbb-inf-it-end}
$\bar{Y} = \{ \{\hat{p}_{j}, \{\hat{y}_{ij}, \bar{\Sigma}_{ij}\}_{i=1}^I \}_{j=1}^J\}$
\end{algorithm}

 The Inference stage of our Double-M Quantification method is shown in Algorithm~\ref{alg:mbb-inf}. For the $i$th corner of the $j$th bounding box, we use $\hat{y}_{ij}$ as the mean of the multivariate Gaussian distribution. we estimate the covariance matrix $\bar{\Sigma}_{ij}$ by using (i) the predicted covariance matrix $\hat{\Sigma}_{ij}$ from the extra regression header and (ii) the estimated aleatoric and epistemic uncertainties $\Sigma_{e}$, $\Sigma_{a}$ obtained from the training stage.  $\bar{\Sigma}_{ij}$ is calculated as the following (see Lines~\ref{alg:mbb-inf-it-start}--\ref{alg:mbb-inf-it-end}):
\begin{equation}
\begin{aligned}
    \bar{\Sigma}_{ij} = \Sigma_{e} + (\frac{1}{2}\Sigma_{a} + \frac{1}{2}\hat{\Sigma}_{ij}).
\end{aligned}\label{eq:cov-comb-output}
\end{equation}

\subsection{Loss Function}
\label{sec:direct_modeling}

In our~\ac{mdm} method, we define the regression loss function of the object detector as the KL-Divergence between the predicted distribution and the ground-truth distribution.

Here we assume all corners are independent, and the distribution of each corner is a multivariate Gaussian distribution:
\begin{equation}
\nonumber
    P_{\theta}(\bar{y}_i|\hat{y}_i,\hat{\Sigma}_i) = \frac{1}{\sqrt{2\pi |\hat{\Sigma}_i|}} \exp{\left(-\frac{(\bar{y}_i - \hat{y}_i)^T \hat{\Sigma}_i^{-1} (\bar{y}_i - \hat{y}_i)}{2} \right)},
\end{equation}
where $\bar{y}_i$ is one possible vector for the $i$-th corner, $\hat{\Sigma}_i$ is a symmetric positive definite $D\times D$ covariance matrix predicted for the $i$-th corner. In the implementation, we utilize the Cholesky decomposition~\cite{kumar2020luvli} to calculate a symmetric positive definite covariance matrix. We compare this distribution with other distributions in Section~\ref{sec:experiment} and demonstrate our selected distribution is the best.

We assume the distribution of each corner for the ground-truth bounding box as a Dirac delta function~\cite{he2019bounding}: \\
   \centerline{$ P_G(\bar{y}_i|y_i) = \delta(\bar{y}_i - y_i).$}

Then, we define the regression loss function for the $i$-th corner as the Kullback–Leibler (KL) divergence between $P_{\theta}(\bar{y}_i|\hat{y}_i,\hat{\Sigma}_i)$ and $P_G(\bar{y}_i|y_i)$~\cite{murphy2012machine}:
\begin{equation}
\begin{aligned}
    \mathcal{L}^i_{KL}(y_i, \hat{y}_i, \hat{\Sigma}_i) = & \frac{1}{2} (y_i - \hat{y}_i)^T \hat{\Sigma}_i^{-1} (y_i - \hat{y}_i) + \\ 
    &  \frac{1}{2} \log |\hat{\Sigma}_i| + \frac{\log(2\pi)}{2} - H(P_G(\bar{y}_i)),
\end{aligned} \label{eq:kl-loss}
\end{equation}
where $H(P_G(\bar{y}_i))$ is the entropy of $P_{G}(\bar{y}_i)$. The last two terms $\frac{\log(2\pi)}{2}$ and $H(P_G(\bar{y}_i))$ could be ignored in the loss function, for they are independent of the model parameters $\theta$. The first term encourages increasing the covariance of the Gaussian distribution as the predicted mean vector diverges from the ground-truth vector. The second regularization term penalizes high covariance to reduce uncertainty.

We add one extra regression header to predict all covariance matrix $\hat{\Sigma}_i$ ($i \in 1,...,I$) with a similar structure as the regression header for $\hat{y}_i$, based on the origin collaborative object detector. 

With a given training dataset, the collaborative object detector $f_{\theta}$ is trained to predict $\hat{Y}$, the classification probability and the regression distribution of  all $J$ objects. The classification loss is $\mathcal{L}_{cls}(c, \hat{p})$ as~\cite{li2022v2x}. The loss function of the object detector is: 
\begin{equation}
    \mathcal{L}_{det} =  \sum_{j=1}^J(\mathcal{L}_{cls}(c_{j}, \hat{p}_{j}) + \sum_{i=1}^I \mathcal{L}^i_{KL}(y_{ij}, \hat{y}_{ij}, \hat{\Sigma}_{ij})).\label{eq:L}
\end{equation}

\section{Experiment}
\label{sec:experiment}

To demonstrate our uncertainty quantification method for collaborative detection, we evaluated it on the V2X-Sim dataset~\cite{li2022v2x} which contains 80 scenes for training, 10 scenes for validation, and 10 scenes for testing. It is generated by CARLA simulation~\cite{Dosovitskiy17}. Each scene contains a 20-second traffic flow at a certain intersection with a 5Hz record frequency, which means each scene contains 100 time-series frames. In each scene, 2-5 vehicles are selected as the connected vehicles and 3D point clouds are collected from LiDAR sensors on them. 

For object detection, we consider Lower-bound, DiscoNet, and Upper-bound for benchmark as follows:
\begin{enumerate}
    \item \ac{lb}~\cite{li2022v2x}: The individual object detector without collaboration which only uses the point cloud data from one individual LiDAR.
    \item \ac{dn}~\cite{li2021learning}: The intermediate collaborative object detector which utilizes a directed graph with matrix-valued edge weight to adaptively aggregate features of different agents by repressing noisy spatial regions while enhancing informative regions. It has shown a good performance-bandwidth trade-off by sharing a compact and context-aware scene representation.
    \item \ac{ub}~\cite{li2022v2x}: The early collaborative object detector uses raw point cloud data from all connected vehicles, as shown in Fig.~\ref{fig:main}(a). It usually has excellent performance with lossless information, yet consumes high communication bandwidth.
\end{enumerate}

\ac{lb} is the traditional individual object detector. \ac{dn} and~\ac{ub} are collaborative object detectors. The basic implementation of all three benchmarks is from the public code of~\cite{li2022v2x}. It uses FaFNet~\cite{luo2018fast}, a classic anchor-based model containing a convolutional encoder, a convolutional decoder, and two output headers for classification and regression, as the backbone of all detectors. 

We compare three uncertainty quantification methods, which are~\ac{dm}, \ac{mbb}, and Double-M Quantification on accuracy and uncertainty. Compared with~\ac{mdm}, the~\ac{mbb} method does not have the predicted covariance matrix during training and inference. 
For~\ac{dm} and~\ac{mdm}, we add one output header for the covariance matrix and the regression loss is the~\ac{kl} loss in Eq~\eqref{eq:kl-loss}. For~\ac{mbb}, the regression loss is the smooth $L_1$ loss. For all, the classification loss is the focal cross-entropy loss~\cite{lin2017focal}.


\begin{table}[t]
  \centering
  \caption{Detection performance of different UQ methods on LB, DN, and UB. Our~\ac{mdm} method improves up to 3.04\% average precision.}
  \vspace{-5pt}
  \label{tab:ap}
  \tabcolsep=4pt
  \begin{tabular}{|c|c|c|c|c|c|c|}
  \hline
   & \multicolumn{3}{c|}{AP @IoU=0.5 $\uparrow$}&\multicolumn{3}{c|}{AP @IoU=0.7 $\uparrow$}\\
  \hline
   UQ Method&\ac{lb}&\ac{dn}&\ac{ub}& \ac{lb}&\ac{dn}&\ac{ub}\\
 \hline
  None & 46.50& 66.60& 69.76& \textbf{41.29}& 60.84&64.78\\
  \hline
  \ac{dm}~\cite{feng2021review}& 43.31& 66.36& 69.28& 39.24& 60.91& 65.25\\
  \hline
  \ac{mbb}~\cite{lahiri1999theoretical}& \textbf{46.64}& 67.10& 70.41& 40.77& 60.98&64.48\\
  \hline
  Double-M (Ours) &43.83 &\textbf{67.20} &\textbf{70.44} &39.74 &\textbf{62.69} &\textbf{66.37}\\
  \hline
  \end{tabular}\\
\end{table}

\subsection{Accuracy Evaluation}
We use~\ac{ap} at~\ac{iou} thresholds of 0.5 and 0.7 as the accuracy measurement, which is widely used in object detection~\cite{feng2021review}. Table~\ref{tab:ap} shows the~\ac{ap} results of different uncertainty quantification (UQ) methods on the test dataset, where ``None" means the object detector is a deterministic one without any UQ method. Our \ac{mdm} achieves better accuracy than others, especially for collaborative object detection. Compared with deterministic detectors without the UQ method, it increases up to 3.04\%~\ac{ap}, which means our proposed~\ac{mdm} method improves the accuracy of collaborative object detection. Meanwhile, DiscoNet and Upper-bound achieve much higher accuracy than Lower-bound  for sharing information between CAVs.

\begin{table}[t]
  \centering
  \caption{NLL comparison of different UQ methods on LB, DN, and UB. Our~\ac{mdm} method achieves up to 4$\times$ improvement on NLL.}
  \vspace{-5pt}
  \label{tab:nll}
  \tabcolsep=4pt
  \begin{tabular}{|c|c|c|c|c|c|c|}
  \hline
   & \multicolumn{3}{c|}{NLL @\ac{iou}=0.5 $\downarrow$}&\multicolumn{3}{c|}{NLL @\ac{iou}=0.7 $\downarrow$}\\
  \hline
  UQ Method&\ac{lb}&\ac{dn}&\ac{ub}& \ac{lb}&\ac{dn}&\ac{ub}\\
  \hline
  \ac{dm}~\cite{feng2021review}& 13.222& 10.015& 14.721& 9.009& 7.896& 13.001\\
  \hline
  \ac{mbb}~\cite{lahiri1999theoretical}& 28.130& 13.794& 22.958& 19.996& 9.710&18.077\\
  \hline
  Double-M (Ours)&\textbf{6.871} &\textbf{5.084} &\textbf{7.974} &\textbf{4.889} &\textbf{3.851} &\textbf{6.696}\\
  \hline
  \end{tabular}\\
\end{table}

\subsection{Uncertainty Evaluation}
\begin{figure*}[t]
    \centering    
    \includegraphics[width=0.95\textwidth]{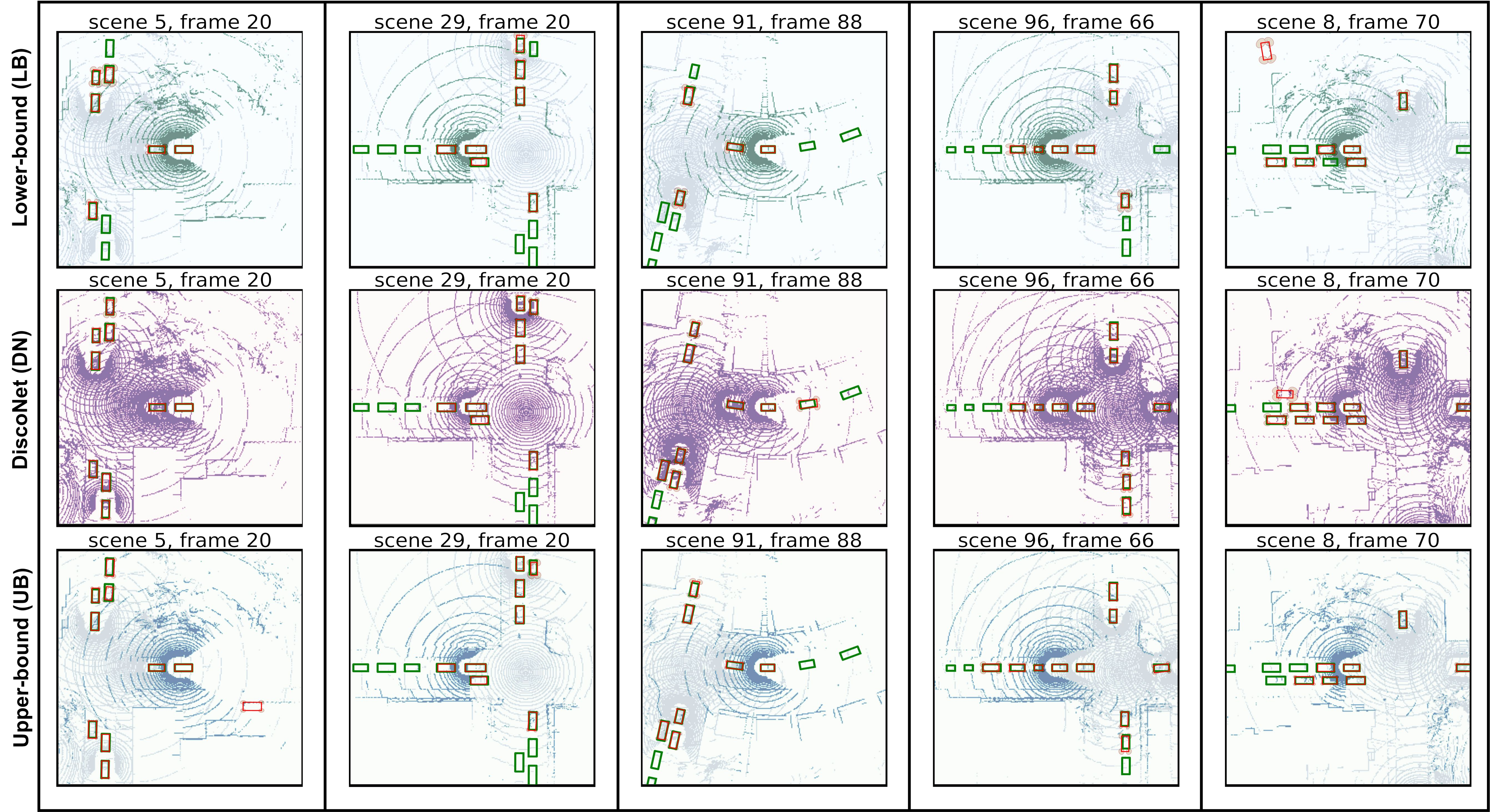}
    \vspace{-8pt}
    \caption{Visualization of our Double-M Quantification  results on different scenes of V2X-Sim~\cite{li2022v2x}. The results of LB, DN, and UB are respectively shown in the first, second, and third rows. Red boxes are predictions, and green
boxes are ground truth. The orange ellipse denotes the covariance of each corner. We can see our Double-M Quantification predicts large orange ellipses when the difference between the red bounding box and the corresponding green bounding box is huge, which means our method is efficient. For example, in the subfigure of DiscoNet on scene 29 frame 20, for the top-right object $O_1$, the difference between the red and green bounding boxes is huge so that~\ac{mdm} predicts large orange ellipses. And for  the three objects on the left side of $O_1$, the difference between their red and green bounding boxes is little so that~\ac{mdm} predicts small orange ellipses.}
    \label{fig:vis_comp}
    \vspace{-10pt}
\end{figure*}

We use~\ac{nll} at~\ac{iou} thresholds of 0.5 and 0.7 as the uncertainty measurement~\cite{feng2021review, harakeh2021estimating}. 
\ac{nll} is a widely used uncertainty score to measure the quality of predicted probability distribution on a test dataset~\cite{meyer2020learning,he2019bounding,lakshminarayanan2017simple,boris2022propagating}. For the test dataset with $K$ frames, it is computed as:\\
  $  NLL = - \frac{1}{I \times J \times K} \sum_{k=1}^{K}\sum_{j=1}^{J}\sum_{i=1}^{I}\log P(y_{ijk}|\hat{y}_{ijk}, \bar{\Sigma}_{ijk}),$
where $\hat{y}_{ijk}$, $\bar{\Sigma}_{ijk}$ are the mean and covariance of the multivariate Gaussian distribution. For~\ac{mbb}, $\bar{\Sigma}_{ijk}$ is $\Sigma_{e}$. For~\ac{dm}, it is the predicted covariance $\hat{\Sigma}_{ijk}$ from the additional regression header. For~\ac{mdm}, it is computed with Eq~\eqref{eq:cov-comb-output}.

Table~\ref{tab:nll} shows the~\ac{nll} results of different uncertainty quantification methods on the test dataset. From Table~\ref{tab:nll}, we can see our proposed \ac{mdm} always achieves much smaller \ac{nll} than other uncertainty quantification methods, with up to 4$\times$ improvement, which means our \ac{mdm} method performs best on uncertainty quantification. Fig.~\ref{fig:vis_comp} shows the qualitative results of our \ac{mdm} method on different scenes. \
For different types of object detectors, DiscoNet always achieves the smallest~\ac{nll}.  
From Table~\ref{tab:nll} and Fig.~\ref{fig:vis_comp}, we can see sharing intermediate feature information between CAVs could improve the uncertainty of object detection.

\subsection{Ablation Study on Uncertainty Distribution}

We compare different probability distributions, which is the key step of the~\ac{dm} method and the~\ac{mdm} method, on accuracy. In particular, we consider the following probability Gaussian distribution:
\begin{enumerate}
    \item Independent Multivariate Gaussian (IMG): Our uncertainty representation of independent Gaussian distribution. All corners are independent, and the distribution of each corner is a multivariate Gaussian distribution.
    \item Independent Single-variate Gaussian (ISG)~\cite{he2019bounding}: Single-variant Gaussian distribution. All corners are independent, and all dimensions of one corner are also independent. We use a single-variate Gaussian distribution for each dimension of each corner. 
    \item Dependent Multivariate Gaussian (DMG)~\cite{he2020deep}: High dimensional Gaussian distribution. All corners are dependent, and the distribution of all corners is a multivariate Gaussian distribution. 
\end{enumerate}

Table~\ref{tab:dis} shows the~\ac{ap} results of ~\ac{dm} method and~\ac{mdm} method for the Upper-bound detector, under different probability distributions. From the table, we can see IMG achieves the best accuracy with up to 4.14\% improvement, which means our uncertainty representation of independent Gaussian distribution for each corner outperforms single-variate Gaussian distribution and high dimensional Gaussian distribution formats. The reason is that our design considers high dependence of all dimensions in one corner and low dependence of all corners.


\begin{table}[t]
  \centering
  \caption{The accuracy comparison of Upper-bound under different probability distributions. Our IMG distribution improves up to 4.14\% average precision.}
  \vspace{-10pt}
  \label{tab:dis}
  \tabcolsep=4pt
  \begin{tabular}{|c|c|c|c|c|}
  \hline
   & \multicolumn{2}{c|}{\ac{ap} @\ac{iou}=0.5 $\uparrow$}&\multicolumn{2}{c|}{\ac{ap} @\ac{iou}=0.7 $\uparrow$} \\
   \hline
   Distribution & \ac{dm} & Double-M &\ac{dm} &Double-M \\
  \hline
  IMG (Ours) & \textbf{69.28} & \textbf{70.44}& \textbf{65.25} & \textbf{66.37} \\
  \hline
  ISG~\cite{he2019bounding} & 68.47&68.95 & 64.68&65.30 \\
  \hline
  DMG~\cite{he2020deep} & 68.23& 67.74&64.86&63.73\\
  \hline
  \end{tabular}
\end{table}

\section{Conclusion}
\label{sec:conclusion}
This work proposes the first attempt to estimate the uncertainty of collaborative object detection. We propose one novel uncertainty quantification method, called Double-M Quantification, to predict both the epistemic and aleatoric uncertainty with one inference pass. The key novelties are the tailored moving block bootstrap training process, and the loss function design that estimates one independent multivariant Gaussian distribution for each corner of the bounding box. We validate our uncertainty quantification method on different collaborative object detectors. Experiments demonstrate that our method achieves better uncertainty estimation and accuracy. In the future, we will apply our method to more collaborative perception datasets, and enhance the performance of trajectory prediction with uncertainty quantification.

\bibliographystyle{IEEEtran}
{ \small 
\balance
\bibliography{uncertainty}
}

\end{document}